\documentclass[
]{ceurart}

\usepackage{listings}
\begin{document}

\copyrightyear{2022}
\copyrightclause{Copyright for this paper by its authors.
  Use permitted under Creative Commons License Attribution 4.0
  International (CC BY 4.0).}

\conference{K-CAP 2025 Posters and Demos: The 13 \textsuperscript{th} International Conference on Knowledge Capture,
  December 10 - 12, 2025, Dayton, Ohio, USA}

\title{A Case Study on Concept Induction for Neuron-Level Interpretability in CNN}


\author[1]{Moumita Sen Sarma}[%
orcid=0009-0003-6644-6531,
email=moumita@ksu.edu,
]
\cormark[1]
\address[1]{Department of Computer Science, Kansas State University, Manhattan, Kansas, USA}

\author[1]{Samatha Ereshi Akkamahadevi}[%
orcid=0009-0001-6333-8004,
email= samatha94@ksu.edu,
]

\author[1]{Pascal Hitzler}[%
orcid=0000-0001-6192-3472,
email=hitzler@ksu.edu,
]
\cormark[1]

\cortext[1]{Corresponding author.}

%
 \begin{abstract}
  Deep Neural Networks (DNNs) have advanced applications in domains such as healthcare, autonomous systems, and scene understanding, yet the internal semantics of their hidden neurons remain poorly understood. Prior work introduced a Concept Induction–based framework for hidden neuron analysis and demonstrated its effectiveness on the ADE20K dataset \cite{dalal2024value}. In this case study, we investigate whether the approach generalizes by applying it to the SUN2012 dataset \cite{xiao2010sun}, a large-scale scene recognition benchmark. Using the same workflow, we assign interpretable semantic labels to neurons and validate them through web-sourced images and statistical testing. Our findings confirm that 
  the method transfers to SUN2012, showing its broader applicability.
\end{abstract}

%
\begin{keywords}
  Explainable AI (XAI) \sep
  Concept Induction \sep
  Hidden Neuron Analysis \sep
  CNNs
\end{keywords}
\maketitle
\vspace{-0.8em}
\section{Introduction}
Deep Neural Networks (DNNs), particularly Convolutional Neural Networks (CNNs), have achieved state-of-the-art performance in image classification and scene understanding. Yet, the hidden neurons remain opaque, limiting interpretability in domains where transparency is critical \cite{angelov2021explainable}. Explainable AI (XAI) techniques such as saliency maps and attribution methods (e.g., SHAP \cite{NIPS2017_7062}, LIME \cite{ribeiro2016should}) highlight input contributions but rarely capture what individual neurons semantically represent.
Concept Induction \cite{LehmannHitzler2010} offers a neurosymbolic alternative by mapping neuron activations to high-level concepts grounded in knowledge graphs, which organize concepts into hierarchies of entities and relations. By contrasting positive and negative activation sets with a structured ontology, it induces logical class expressions for each neuron that yield candidate semantic labels.
In previous work \cite{dalal2024value}, this approach was applied to the ADE20K dataset \cite{zhou2019semantic} with the automation pipeline in \cite{akkamahadevi2024automating}, demonstrating that neurons could be systematically assigned interpretable labels with strong empirical support. Herein, we apply the same methodology to a different large-scale benchmark, SUN2012 \cite{xiao2010sun}.
Our objective is to assess whether the findings extend beyond ADE20K. By validating the approach on SUN2012, we indeed show that the method transfers and that robust neuron–concept associations emerge consistently across datasets.

\vspace{-0.8em}
\section{Methodology}
The hidden neuron analysis follows a structured workflow that connects hidden neuron activations to human-understandable concepts. We follow the approach from \cite{dalal2024value}, consisting of data preparation, model training, activation extraction, concept induction, and evaluation of the induced concepts.
\textbf{Data Selection \& Preparation:}
    SUN2012 contains 131,000 images across 908 scene categories with annotations for over 3,800 objects. For this study, the ten largest categories are selected: bathroom, bedroom, building facade, dining room, highway, kitchen, living room, mountain snowy, skyscraper, and street, yielding 3,157 images for training and validation and 793 for testing.  \\   
\textbf{Model Training:}
As in \cite{dalal2024value}, multiple CNN architectures (VGG16/19, InceptionV3, ResNet50/101/152/50V2) are fine-tuned. The models were trained using their standard input resolutions: 299 × 299 for InceptionV3 and 224 × 224 for the remaining architectures. Training was carried out with the Adam optimizer set to a learning rate of 0.001, using categorical cross-entropy as the loss function. A batch size of 32 was used, and training ran for up to 30 epochs. To prevent overfitting, early stopping was applied by monitoring the validation loss with a patience of three epochs, restoring the best-performing model weights.While the prior study on ADE20K reported ResNet50V2 as the best-performing model, in our experiments on SUN2012, InceptionV3 achieves the highest performance (96.83\% training accuracy; 92.71\% validation accuracy) and is therefore selected for further analysis.\\
 \textbf{Neuron Activation Extraction:}
 Following the process introduced in previous work \cite{dalal2024value}, neuron activations are extracted from the dense layer of the trained network. Each test image is passed through the network, and activation values of all 64 dense-layer neurons are collected.
 To capture activation behavior, thresholds are defined relative to the maximum observed response: images at or above 80\% form the positive set, while those at or below 20\% form the negative set. These contrasting sets are then used in the Concept Induction stage to derive the semantic labels.\\
\textbf{Concept Induction:}
The Efficient Concept Induction and Integration (ECII) system \cite{SarkerHitzler2019} is employed to derive concepts from the neuron activation sets. For every image, a minimal ontology is constructed by considering only the annotated objects and mapping them to exact lexical matches within a Wikipedia-based concept hierarchy  \cite{DBLP:conf/kgswc/SarkerSHZNMJRA20}. These image-specific ontologies are then integrated with the Wikipedia hierarchy through ECII to form a single, robust background ontology. ECII then  made use of this background knowledge to generate logical class expressions evaluated using coverage score that distinguishes positive from negative sets and that is defined by
\vspace{-0.8em}
\begin{equation}
\text{coverage}(E) = \frac{|Z_1| + |Z_2|}{|P \cup N|},\nonumber
\end{equation}
\noindent
where \(Z_1 = \{\, p \in P \mid K \models E(p) \,\}\) and
\(Z_2 = \{\, n \in N \mid K \not\models E(n) \,\}\), \(P\) and  \(N\) refer to positive set and negative set, and \(K\) represents the knowledge base. Coverage measures how well the induced concept aligns with a neuron’s activation pattern, with higher scores indicating a better match.
This stage resulted in candidate semantic labels (e.g., snowy mountain, toilet tissue, crosswalk) for neurons.\\
\textbf{Concept Evaluation:}
We adopt the same evaluation procedure as in the ADE20K study, combining web-sourced image confirmation with statistical testing.
For label confirmation of each neuron, up to 100 Google Images are retrieved and tested with 80\% of the retrieved images. Label of a neuron is confirmed when Target Level Activation (TLA) $\geq$ 80\%, a measurement of the proportion of images associated with a concept that reliably activates the neuron. On the other hand, Non-Target Label Activation (Non-TLA) is defined as the proportion of those same images in which a different neuron reliably activates above the threshold.
%
For statistical validation, Mann–Whitney U test, a non-parametric statistical test is performed on 20\% of the retrieved images. 
Here, we require p < 0.05 with a negative z-score, which demonstrates that target images consistently trigger stronger activations than non-target images. This results in rejecting the null hypothesis, which states that there is no significant difference between activations for target images (retrieved with the neuron’s label) and non-target images (retrieved with other labels).

\vspace{-0.8em}
\section{Results and Discussion}
\vspace{-0.8em}
The application of Concept Induction to the SUN2012 dataset produces a set of interpretable neuron labels that aligned strongly with semantic concepts. Out of the 64 dense-layer neurons analyzed, 32 are confirmed to exhibit stable concept associations with Target Label Activations (TLA) of at least 80\%.
Of these 32 neurons, 29 neurons also show statistically significant separation between target and non-target activations under the Mann–Whitney U test (p < 0.05), confirming that their responses were meaningfully stronger for concept-related images. Table \ref{neuron_stats} shows the result of statistical test, where confirmed labels include crosswalk, skyscraper, pillow, ceiling fan, and bidet, 


Compared to ADE20K, which yielded 19 confirmed neurons, SUN2012 produces 32 under the same evaluation procedure. This shows that despite dataset and architectural differences (ResNet50V2 for ADE20K and InceptionV3 for SUN2012), the Concept Induction framework reliably identifies semantically coherent neurons across benchmarks. Therefore, it provides fine-grained, human-readable, and verifiable neuron-level explanations, supporting transparent analysis, greater trust, and practical debugging of deep models.

\vspace{-0.8em}

\begin{table}[!t]
\centering
\footnotesize
\caption{Label confirmation via Google Images results with confirmed neurons (TLA \% $\geq$ 80).}
\label{neuron_google}
\begin{tabular}{|c|l|c|c|c|}
\hline
	\textbf{Neuron ID} & \textbf{ECII Concepts} & \textbf{Coverage Score} & \textbf{TLA \%} & \textbf{Non-TLA \%} \\
\hline\hline
0  & snowy\_mountain & 0.986 & 95.00 & 52.57 \\
7  & snow, mountain & 0.998 & 80.00 & 39.55 \\
9  & field & 0.971 & 93.75 & 74.96 \\
13 & dish\_rack & 0.892 & 81.25 & 31.12 \\
15 & pillow, ceiling\_fan & 0.992 & 88.75 & 57.15 \\
16 & skyscraper, river\_water & 0.997 & 80.00 & 42.05 \\
18 & city & 0.993 & 87.50 & 68.62 \\
19 & snowy\_mountain & 0.975 & 97.50 & 46.58 \\
21 & bedroom, duvet & 0.967 & 85.00 & 47.67 \\
22 & dishwasher & 0.945 & 96.20 & 24.58 \\
23 & fence & 0.965 & 98.75 & 77.85 \\
24 & sink & 0.939 & 96.25 & 52.59 \\
26 & toilet\_tissue & 0.979 & 95.00 & 39.49 \\
27 & toilet & 0.961 & 95.00 & 55.33 \\
28 & cars & 0.903 & 97.50 & 72.79 \\
31 & snowy\_mountain & 0.948 & 97.50 & 67.00 \\
32 & shell & 0.950 & 80.00 & 46.64 \\
33 & bed, pillow & 0.945 & 80.00 & 34.41 \\
36 & snowy\_mountain & 0.973 & 98.75 & 68.33 \\
40 & plant & 0.972 & 83.75 & 51.00 \\
41 & pole, handrail & 0.997 & 85.00 & 70.35 \\
42 & skyscraper & 0.969 & 93.75 & 49.47 \\
43 & skyscraper & 0.978 & 100.00 & 66.88 \\
47 & crosswalk & 0.983 & 81.25 & 23.42 \\
48 & bidet & 0.966 & 98.75 & 57.35 \\
49 & ironing\_board, shoe & 0.969 & 93.75 & 63.09 \\
50 & field & 0.976 & 83.75 & 66.54 \\
54 & snowy\_mountain & 0.965 & 92.50 & 40.80 \\
58 & snowy\_mountain & 0.969 & 97.50 & 55.05 \\
60 & brocoli, cheese & 0.979 & 93.75 & 39.00 \\
61 & cars & 0.976 & 90.00 & 45.51 \\
62 & air\_conditioning, chest\_of\_drawers & 0.982 & 87.50 & 61.18 \\
\hline
\end{tabular}
\end{table}

\begin{table}[!h]
\large
\centering
\caption{Statistical evaluation results. \textbf{Bold} rows represents neurons with p-value $\geq$ 0.05, where we \textbf{cannot reject the null hypothesis}.}
\label{neuron_stats}
\resizebox{\columnwidth}{!}{%
\begin{tabular}{
|>{\centering\arraybackslash}p{0.09\linewidth}
|>{\centering\arraybackslash}p{0.41\linewidth}
|>{\centering\arraybackslash}p{0.07\linewidth}
|>{\centering\arraybackslash}p{0.12\linewidth}
|>{\centering\arraybackslash}p{0.10\linewidth}
|>{\centering\arraybackslash}p{0.10\linewidth}
|>{\centering\arraybackslash}p{0.10\linewidth}
|>{\centering\arraybackslash}p{0.15\linewidth}
|>{\centering\arraybackslash}p{0.11\linewidth}
|>{\centering\arraybackslash}p{0.14\linewidth}
|
}

\hline
	\textbf{Neuron ID} & \textbf{ECII Concepts} & \textbf{TLA \%} & \textbf{Non-TLA \%} & \textbf{Target Median} & \textbf{Non-Target Median} & \textbf{Target Mean} & \textbf{Non-Target Mean} & \textbf{$z$-score} & \textbf{$p$-value} \\
\hline\hline
0 & snowy\_mountain & 95 & 53.02 & 5.99 & 0.23 & 5.58 & 1.22 & -6.18 & $<0.00001$ \\
\hline
7 & snow, mountain & 70 & 39.68 & 2.10 & 0.00 & 1.74 & 0.68 & -3.04 & $0.00061$ \\
\hline
9 & field & 100 & 75.48 & 4.38 & 1.43 & 4.03 & 1.97 & -3.64 & $0.00025$ \\
\hline
13 & dish\_rack & 90 & 32.62 & 1.65 & 0.00 & 1.68 & 0.56 & -4.66 & $<0.00001$ \\
\hline
15 & pillow, ceiling\_fan & 90 & 54.21 & 3.03 & 0.21 & 2.36 & 0.85 & -4.40 & $<0.00001$ \\
\hline
16 & skyscraper, river\_water & 70 & 43.41 & 1.54 & 0.00 & 1.44 & 0.45 & -3.15 & $0.00052$ \\
\hline
18 & city & 80 & 71.11 & 2.17 & 0.82 & 2.03 & 1.10 & -2.85 & $0.00398$ \\
\hline
19 & snowy\_mountain & 100 & 45.95 & 5.91 & 0.00 & 5.39 & 1.09 & -6.55 & $<0.00001$ \\
\hline
21 & bedroom, duvet & 65 & 45.95 & 0.86 & 0.00 & 1.91 & 0.57 & -2.70 & $0.00332$ \\
\hline
22 & dishwasher & 80 & 26.11 & 1.60 & 0.00 & 1.87 & 0.29 & -5.04 & $<0.00001$ \\
\hline
23 & fence & 100 & 78.73 & 3.07 & 1.61 & 2.96 & 1.80 & -3.40 & $0.00063$ \\
\hline
24 & sink & 95 & 53.25 & 3.14 & 0.11 & 3.14 & 0.81 & -5.60 & $<0.00001$ \\
\hline
26 & toilet\_tissue & 100 & 38.65 & 2.47 & 0.00 & 2.36 & 0.51 & -6.29 & $<0.00001$ \\
\hline
27 & toilet & 95 & 56.98 & 4.06 & 0.33 & 3.88 & 0.94 & -6.10 & $<0.00001$ \\
\hline
\textbf{28} & \textbf{cars} & \textbf{95} & \textbf{72.78} & \textbf{1.58} & \textbf{1.03} & \textbf{1.66} & \textbf{1.59} & \textbf{-1.10} & $\mathbf{0.26788}$ \\
\hline
31 & snowy\_mountain & 90 & 67.62 & 4.58 & 0.62 & 3.97 & 1.35 & -5.05 & $<0.00001$ \\
\hline
32 & shell & 85 & 46.19 & 1.65 & 0.00 & 1.55 & 0.70 & -3.77 & $0.00004$ \\
\hline
33 & bed, pillow & 75 & 35.87 & 3.19 & 0.00 & 2.60 & 0.43 & -4.62 & $<0.00001$ \\
\hline
36 & snowy\_mountain & 100 & 67.62 & 4.72 & 0.93 & 4.60 & 1.48 & -6.24 & $<0.00001$ \\
\hline
\textbf{40} & \textbf{plant} & \textbf{70} & \textbf{51.90} & \textbf{1.00} & \textbf{0.07} & \textbf{0.86} & \textbf{0.68} & \textbf{-1.44} & $\mathbf{0.12808}$ \\
\hline
41 & pole, handrail & 90 & 69.21 & 2.23 & 1.05 & 2.05 & 1.51 & -2.18 & $0.02735$ \\
\hline
42 & skyscraper & 95 & 50.79 & 2.91 & 0.05 & 2.78 & 1.01 & -4.49 & $<0.00001$ \\
\hline
43 & skyscraper & 100 & 66.51 & 2.94 & 0.81 & 3.01 & 1.10 & -5.65 & $<0.00001$ \\
\hline
47 & crosswalk & 95 & 22.94 & 3.20 & 0.00 & 3.20 & 0.29 & -6.74 & $<0.00001$ \\
\hline
48 & bidet & 100 & 58.02 & 3.00 & 0.43 & 2.91 & 1.01 & -5.35 & $<0.00001$ \\
\hline
49 & ironing\_board, shoe & 90 & 62.06 & 1.98 & 0.55 & 2.20 & 1.02 & -3.37 & $0.00053$ \\
\hline
\textbf{50} & \textbf{field} & \textbf{85} & \textbf{66.83} & \textbf{0.98} & \textbf{0.73} & \textbf{1.15} & \textbf{1.11} & \textbf{-0.89} & $\mathbf{0.36257}$ \\
\hline
54 & snowy\_mountain & 100 & 39.68 & 4.53 & 0.00 & 4.01 & 0.72 & -6.58 & $<0.00001$ \\
\hline
58 & snowy\_mountain & 95 & 52.94 & 4.56 & 0.18 & 4.77 & 1.12 & -6.33 & $<0.00001$ \\
\hline
60 & brocoli, cheese & 95 & 39.52 & 1.79 & 0.00 & 1.93 & 0.50 & -5.89 & $<0.00001$ \\
\hline
61 & cars & 85 & 45.95 & 1.41 & 0.00 & 1.32 & 0.55 & -4.07 & $<0.00001$ \\
\hline
62 & air\_conditioning, chest\_of\_drawers & 85 & 62.22 & 2.12 & 0.50 & 2.59 & 0.95 & -3.92 & $0.00006$ \\
\hline
\end{tabular}
}
\end{table}

\begin{acknowledgments}
\vspace{-0.8em}
The authors acknowledge partial funding through the Kansas State University Game-changing Research Initiation Program (GRIP). 
\end{acknowledgments}

\vspace{-0.8em}
\section*{Declaration on Generative AI}
\vspace{-0.8em}
  
 For this work, the authors used ChatGPT-5  for grammar and spelling checks. Subsequently, the authors reviewed and edited the content as needed and take full responsibility for the content.

\vspace{-0.8em}
\bibliography{ref}

\appendix



\end{document}